# mHC-SSM: Manifold-Constrained Hyper-Connections for State Space Language Models with Stream-Specialized Adapters

Abdulvahap Mutlu*[1], Şengül Doğan[1], Türker Tuncer[1]

[1]Department of Digital Forensics Engineering, Technology Faculty, Firat University, Elazig, Turkey

241144107@firat.edu.tr; sdogan@firat.edu.tr; turkertuncer@firat.edu.tr

**Abstract:**

Manifold-Constrained Hyper-Connections (mHC) introduce a stability-motivated variant of multi-stream residual mixing by constraining residual stream mixing matrices to the manifold of doubly stochastic matrices via Sinkhorn-Knopp projection. In this work, we study whether mHC-style constrained multi-stream residual topology transfers effectively to state space model (SSM) language modeling. We implement a static mHC mechanism around an SSM block by expanding the residual stream into multiple parallel streams, aggregating streams into a single SSM input through simplex-constrained pre-mixing, scattering the SSM output back to streams through simplex-constrained post-mixing, and applying Sinkhorn-projected residual stream mixing at each layer. We further introduce stream-specialized adapters that add lightweight stream-specific capacity through a shared bottleneck with per-stream scaling, applied both before stream aggregation and after the SSM output prior to scattering. We evaluate baseline single-stream SSM, static mHC SSM, and mHC SSM with adapters on WikiText-2 using identical training settings and report checkpoint-based validation loss, perplexity, throughput, and peak GPU memory. Under the reported fair checkpoint evaluation, static mHC improves validation loss from 6.3507 to 6.2448 and reduces perplexity from 572.91 to 515.35, while mHC with adapters further improves validation loss to 6.1353 and perplexity to 461.88. These gains are accompanied by modest throughput reductions from 1025.52 to 964.81 and 938.90 tokens per second, and increased peak memory from 2365 MB to 2568 MB and 3092 MB. The results suggest that mHC-inspired constrained multi-stream residual mixing can yield measurable quality improvements in SSM language models and that stream-specialized adapter capacity can further enhance performance with predictable efficiency tradeoffs.

## 1. Introduction

Residual connections remain one of the most influential architectural primitives in modern deep learning because they provide a direct path that supports stable optimization as networks become deeper[1,2]. The identity mapping component of the residual formulation has been analyzed as an important factor behind the forward and backward signal propagation properties of residual networks. In contemporary large language models (LLMs), residual connections remain ubiquitous, even as the residual function has evolved to include a wide range of operations such as attention mechanisms and feed-forward sublayers[3].

Despite the success of the residual paradigm, there is ongoing interest in macro-architectural designs that increase model expressivity by modifying how information flows across depth. Hyper-Connections (HC) is one such proposal that expands the residual stream into multiple parallel streams, and introduces learnable mappings that mix and route information between these streams without changing the core layer function's floating point operation count in the same way that widening the inner block would[4]. However, the mHC paper argues that the unconstrained residual stream mixing in HC can compromise the identity mapping property when applied recursively across many layers, and that this can lead to instability due to unbounded amplification or attenuation of signals in forward and backward propagation[5].

Manifold-Constrained Hyper-Connections (mHC) is introduced as a response to these issues. The key idea is to constrain the residual mixing mapping by projecting it onto the manifold of doubly stochastic matrices, also described as the Birkhoff polytope, using the Sinkhorn-Knopp algorithm[6]. The mHC paper motivates the use of doubly stochastic residual mappings by emphasizing several stability-oriented properties, including a spectral norm bound and closure under multiplication, together implying non-expansive behavior and stable composition across depth[7].

This work studies mHC-style constrained multi-stream residual mixing in a non-attention sequence model built from a state space modeling (SSM) block. SSM-based sequence modeling has re-emerged as a competitive alternative backbone for long-sequence modeling due to favorable scaling properties and growing evidence of strong performance when paired with modern architectural and kernel optimizations[8,9]. We implement static mHC-style residual stream mixing inside a lightweight SSM language model and evaluate it on WikiText-2[10]. We further introduce stream-specialized adapters that add compact stream-specific capacity via a shared bottleneck with per-stream scaling, applied both before and after the SSM core. The evaluation includes validation loss and perplexity as well as throughput and peak GPU memory measurements computed under a checkpoint-based, weight-restoring benchmarking protocol.

### 1.1 Literature Review

The residual connection paradigm can be summarized as learning an additive residual function around an identity path, which empirically and theoretically supports deeper networks by enabling direct signal propagation. Related residual-stability studies further show that modifying or preserving residual pathways can improve optimization in very deep networks. Stochastic Depth trains residual networks by randomly bypassing layers with identity mappings, while ReZero gates residual branches with zero-initialized parameters to improve signal propagation at large depth[11,12]. In the sequence modeling domain, the Transformer architecture popularized a design in which residual connections wrap attention and feed-forward sublayers, and that approach has become a widely adopted default in LLM pretraining. The mHC paper frames this broader development as a continued reliance on the original residual form, even as the residual function has diversified in content.

Beyond attention-based architectures, state space models provide an alternative approach to sequence modeling. Structured state space models such as S4 introduced parameterizations and computational strategies that make long-range recurrence efficient, and later selective state space models such as Mamba argue for input-dependent mechanisms that improve performance on information-dense modalities such as language while maintaining linear-time sequence modeling characteristics. While this work does not claim equivalence to those specific architectures, they establish that SSM-based backbones are an active and relevant design space for language modeling, motivating investigations of macro-architectural changes such as multi-stream residual topologies. Additional evidence comes from diagonal and language-focused SSM variants: DSS shows that diagonal state-space parameterizations can be competitive with more structured SSMs, while H3 directly studies language modeling with SSM layers and identifies recall and token-comparison as key challenges for closing the gap with attention[13,14].

Hyper-Connections proposes to expand the residual stream width and to introduce learned pre, post, and residual mappings that aggregate from multiple streams into the layer input, scatter the layer output back to streams, and mix within the residual stream. The mHC paper highlights that when such unconstrained residual mixing is composed across many layers, the resulting composite mapping can fail to preserve the global mean of features across streams, potentially producing amplification or attenuation that leads to training instability. To address this, mHC constrains the residual mixing to be doubly stochastic, meaning that entries are non-negative and each row and column sums to one.

The choice of doubly stochastic residual mixing is motivated by properties described directly in the mHC paper. It claims that a doubly stochastic matrix has spectral norm bounded by one, that the set is closed under multiplication, and that the manifold can be interpreted as the convex hull of permutation matrices, yielding a geometric view of stable, progressive mixing. The paper further discusses non-negativity constraints on the input and output mappings to avoid signal cancellation effects from mixtures of positive and negative coefficients.

Finally, adapters and other parameter-efficient mechanisms introduce additional capacity via small inserted modules, frequently implemented as a down projection to a low-dimensional bottleneck followed by an up projection back to the model dimension. Adapter modules were proposed as a way to add compact task-specific capacity while preserving most of a pretrained model's parameters[15]. Low-rank adaptation methods such as LoRA provide related parameter-efficient strategies based on low-rank decompositions inserted into linear layers[16]. Although our adapters are not used for task adaptation in a pretrained setting, these works motivate the general principle that compact bottleneck mechanisms can add meaningful capacity.

### 1.2 Literature Gaps

The mHC paper introduces and motivates manifold-constrained stream mixing, emphasizing stability and scalability in the context of large-scale language model training, and it explicitly pairs the method with system-level optimizations to address the memory access overhead induced by widened residual streams. However, the behavior of mHC-style residual mixing in SSM-based language models is not established by the paper itself, and the interaction between constrained stream mixing and SSM-style recurrence is not obvious a priori.

In addition, while mHC constrains mixing to preserve stability-oriented properties, it does not directly address how one might introduce explicit stream specialization capacity on top of constrained mixing in a lightweight way, particularly in settings where the layer function is a compact SSM block and the goal is to improve perplexity without extensively increasing the core block size. This suggests an opportunity to study whether stream-specific capacity can complement constrained mixing and whether the resulting tradeoffs in throughput and memory align with practitioner constraints.

### 1.3 Motivation

SSM-based language models represent an active alternative to attention-based architectures, often motivated by efficiency considerations[17]. Yet, like other deep models, SSM stacks are subject to optimization challenges as depth increases, and performance improvements can depend on macro-architectural choices as much as micro-layer design. Hyper-Connections and mHC propose a macro-architectural mechanism to increase topological expressivity through multiple residual streams.

This work is motivated by the question of whether the stability-oriented constraints of mHC transfer beneficially to SSM language models, and whether adding small stream-specialized adapter capacity can yield further quality improvements while keeping the SSM core unchanged.

A second motivation is practical. Multi-stream residual designs introduce compute and, importantly, memory access overhead. The mHC paper explicitly highlights that although HC can preserve floating point operation counts, the memory access costs of widened residual streams are a key systems concern, and it proposes infrastructure-level optimizations such as kernel fusion and custom backward computation to address those bottlenecks.

In many research settings, however, implementations are conducted in general-purpose frameworks without specialized fused kernels. It is therefore useful to quantify how quality improvements trade off against throughput and peak memory in a straightforward PyTorch implementation.

### 1.4 Innovation & Contributions

This work makes three contributions grounded in the implementation and experimental logs provided in this chat.

- First, it instantiates static mHC-style residual stream mixing in an SSM language model, using a Sinkhorn-style projection of a learned residual mixing matrix and simplex-constrained pre and post stream mixing weights, and evaluates this variant against a baseline single-stream SSM language model.
- Second, it introduces stream-specialized adapters designed specifically for a multi-stream SSM architecture. The adapters share a bottleneck down projection and up projection across streams, and introduce stream-specific scaling parameters in the bottleneck for both a pre-adapter applied to stream tensors and a post-adapter applied to the SSM output prior to stream scattering. This design adds stream-specific degrees of freedom without altering the SSM recurrence itself.
- Third, it reports comparative results for baseline SSM, static mHC SSM, and mHC plus adapters under identical training settings on WikiText-2, including stepwise validation loss and perplexity logs during training, and a checkpoint-based fair benchmarking report that includes validation loss, perplexity, tokens per second, and peak GPU memory.

## 2. Methods

### 2.1 Task Definition and Dataset

We consider next-token prediction on WikiText-2, a widely used language modeling dataset introduced alongside the WikiText benchmark suite. WikiText datasets preserve punctuation and casing and provide standardized train, validation, and test splits.

Tokenization is performed using a GPT-2 tokenizer through the Hugging Face Transformers ecosystem[18]. The GPT-2 configuration documentation states a default vocabulary size of 50,257. In our implementation, if the tokenizer lacks an explicit pad token, the end-of-sequence token is used as padding.

The dataset is converted into a contiguous token sequence for training and validation. Training samples are created by packing this sequence into fixed-length segments of length $T$, producing an input token sequence $x$ and a target token sequence $y$ that is shifted by one token. This yields the standard autoregressive language modeling objective.

## 2.2 Baseline Single-Stream SSM Language Model

The baseline is a custom lightweight diagonal SSM language model inspired by modern state-space sequence modeling. It uses a residual stack of diagonal SSM blocks rather than the full S4, DSS, or Mamba architectures.

### 2.2.1 Notation

Let $B$ denote batch size, $T$ denote sequence length, $V$ denote vocabulary size and $D$ denote model dimension. The input token indices are $x \in \{0, \dots, V-1\}^{BxT}$ and targets are $y \in \{0, \dots, V-1\}^{BxT}$. The baseline model maps tokens to embeddings and applies learned positional encodings. Let $E \in R^{VxD}$ denote the token embedding matrix, and let $P \in R^{TxD}$ denote learned positional embeddings. The initial hidden states are

$$h_0 = E[x] + P \tag{1}$$

Followed by dropout.

### 2.2.2 Residual SSM Block Structure

The baseline network consists of $\boldsymbol{L}$ identical residual blocks with a state space modeling unit. Each block implements the standard residual update

$$h_{l+1} = h_l + SSMBlock(h_l), l = 0, \dots, L-1 \tag{2}$$

This matches the residual paradigm described in the mHC paper in which the identity mapping $h_l$ is passed directly to depth $l + 1$ and residual functions accumulate additively across depth.

Each SSMBlock includes normalization, a gated linear projection, a causal depthwise convolution, a pointwise nonlinearity, a diagonal state space recurrence computed by scanning over time, and an output projection. For normalization, RMSNorm is used, a variant that normalizes by the root mean square statistic without re-centering[19].

Let $Norm(\cdot)$ denote RMSNorm. Let the input be $h \in R^{BxTxD}$. The block computes

$$\tilde{h} = Norm(h) \tag{3}$$

Then applies a linear projection to 2D channels and splits into an input path and a gate:

$$(u, g) = Linear_{2D}(\tilde{h}), u, g \in R^{BxDxT} \tag{4}$$

A causal depthwise convolution is applied over the time dimension to $u$. The convolution is implemented with left padding of $k - 1$ and cropping to preserve sequence length $T$. After convolution, a SiLU nonlinearity is applied[20]:

$$u' = SiLU(DWConv(u)) \tag{5}$$

The resulting sequence $u'$is passed through a diagonal state space model, described next.

**2.2.3 Diagonal State Space Recurrence**

The diagonal SSM is defined as a per-dimension recurrence:

$$s_t = a \odot s_{t-1} + b \odot u'_t \tag{6}$$

$$z_t = c \odot s_t + d \odot u'_t \tag{7}$$

Where $s_t, u'_t, z_t \in R^{BxD}$, and $a, b, c, d \in R^D$ are learned parameters. Stability is encouraged by constraining $a$ elementwise into (0,1) using a sigmoid parameterization and clamping away from the endpoints. The recurrence is computed in float32 for numerical robustness under automatic mixed precision, and the output is cast back to the input dtype.

Finally, the recurrence output is gated and projected:

$$\hat{z_t} = z_t \odot \sigma(g_t) \tag{8}$$

$$SSMBlock(h) = Linear_D(\hat{z}) \tag{9}$$

With dropout applied if configured.

#### 2.2.4 Output and Loss

After $L$ blocks, a final normalization is applied and a tied linear language modeling head produces logits

$$\ell = Linear_V(Norm(h_L)) \tag{10}$$

Cross-entropy loss is computed between $\ell$ and target tokens $y$ across all positions. Perplexity is computed as $\exp(val_loss)$, a standard conversion used in language modeling.

### 2.3 Hyper-Connections and mHC Background

The mHC paper introduces the HC formulation by expanding the residual stream width by a factor $n$. For the $l$-th layer, HC defines propagation as

$$x_{l+1} = H_l^{res} x_l + \left(H_l^{post}\right)^{\top} F\left(H_l^{pre} x_l, W_l\right) \tag{11}$$

Where $x_l$ and $x_{l+1}$are in $R^{nC}$,$H_l^{res} \in R^{nxn}$ mixes features within the residual stream, and $H_l^{pre}$, $H_l^{post} \in R^{1xn}$ aggregate and scatter between the $nC$-dimensional stream representation and the $C$-dimensional layer input and output.

The paper argues that unconstrained HC can break identity-mapping-like stability as depth increases, because the composite mapping formed by multiplying unconstrained $H^{res}$ matrices does not preserve global mean intensity across streams and can yield unbounded amplification or attenuation.

To address this, mHC constrains $H_l^{res}$ s to be doubly stochastic by projecting it onto the Birkhoff polytope. The doubly stochastic constraint is stated as non-negativity with row and column sums equal to one. The paper motivates this choice by stating that doubly stochastic matrices have spectral norm bounded by one, are closed under multiplication, and form a convex hull of permutation matrices. The paper additionally states that it imposes non-negativity constraints on pre and post mappings to prevent signal cancellation.

mHC parameterizes mappings and then applies manifold projections. In one formulation shown in the paper, it applies a sigmoid nonlinearity to $H^{pre}$ and a scaled sigmoid to $H^{post}$, and applies Sinkhorn-Knopp to $H^{res}$. The Sinkhorn-Knopp procedure is described as exponentiating entries to make them positive and then alternately normalizing rows and columns. The paper states that it uses $t_{max} = 20$ iterations as a practical value in its experiments.

The paper also highlights that system overhead is a central concern, and describes infrastructure optimizations such as kernel fusion, mixed-precision strategies, fusing operations to reduce memory bandwidth bottlenecks, implementing Sinkhorn iterations within a single kernel, and deriving custom backward kernels with recomputation. These points matter for interpreting our own throughput and memory results, since our implementation does not incorporate those specialized kernels.

## 2.4 Static mHC-Style Multi-Stream SSM Architecture

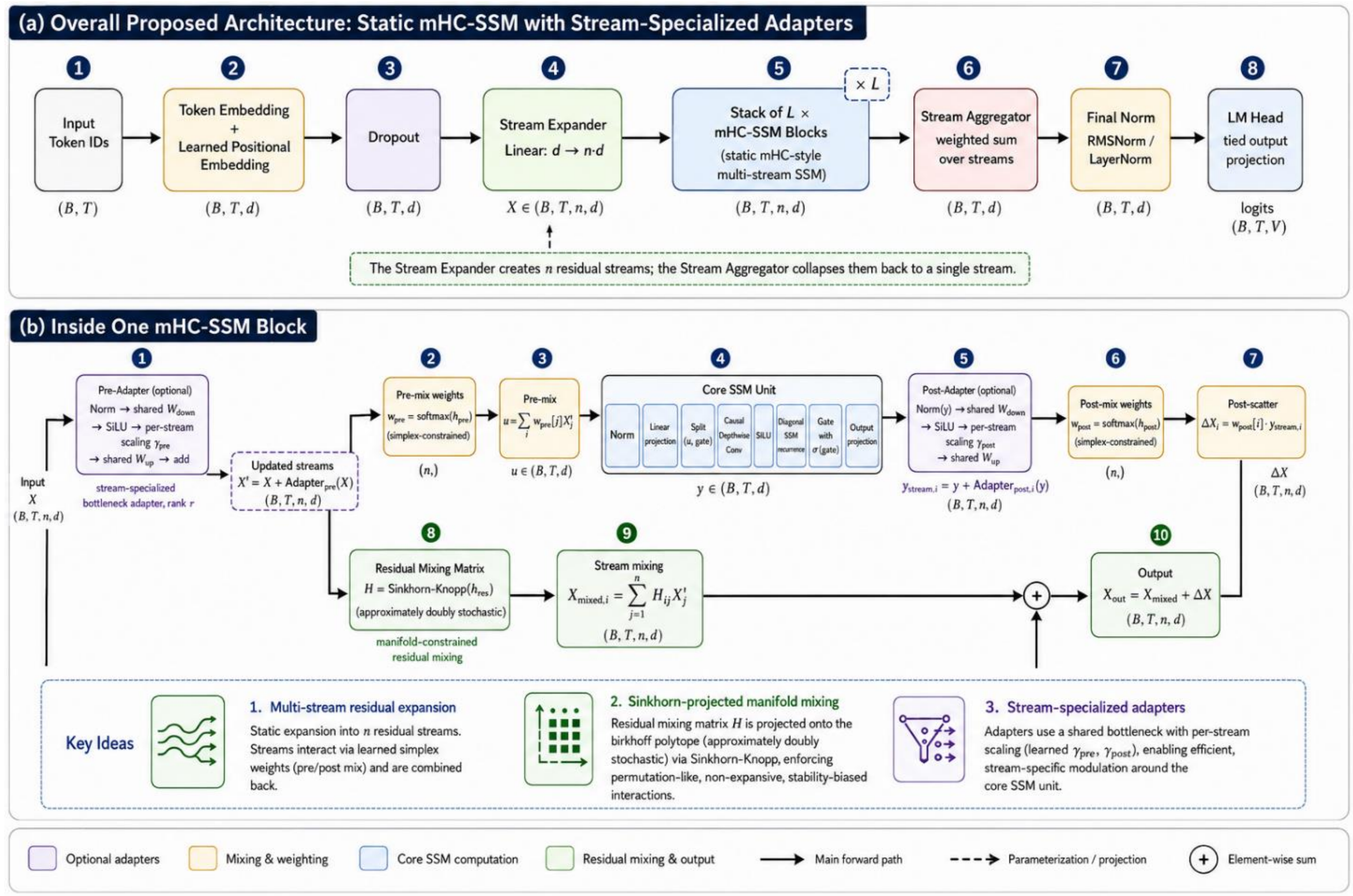

Figure 1. Proposed mHC-SLM architecture and internal mHC-SSM block design.

(a) The full language modeling pipeline expands token representations into multiple residual streams, processes them through a stack of static mHC-SSM blocks, aggregates streams, and produces logits through a tied language modeling head.

(b) Each mHC-SSM block applies optional stream-specialized adapters, simplex-constrained pre/post stream mixing, a compact SSM core, Sinkhorn-projected residual stream mixing, and residual update aggregation.

### 2.4.1 Stream Expansion and Aggregation

In our implementation (as seen in Figure 1), the baseline hidden state is expanded into $n$ streams using a learned linear projection. Let $h \in R^{BxTxD}$ be the single-stream hidden state. A stream expander produces

$$X = Expand(h) \in R^{BxTxnxD}. \tag{12}$$

This provides a multi-stream representation analogous to the expanded residual stream in HC.

To feed an SSM block that operates on a single stream, we aggregate streams using a learned simplex weight vector. Let $w^{pre} \in \Delta^{n-1}$ denote a probability simplex vector. In practice, we parameterize $w^{pre}$ using logits and a softmax, ensuring non-negativity and that weights sum to one. The pre-mixed input is

$$u = \sum_{i=1}^{n} w_i^{pre} X_i \quad (13)$$

Where $X_i \in R^{BxTxD}$ denotes the $i$-th stream slice. The SSM block is applied to $u$ to produce $y \in R^{BxTxD}$. We then scatter the output back to streams using a post-mix simplex weight vector $w^{post} \in \Delta^{n-1}$. In the simplest case, the scattered update is

$$\Delta X_i = w_i^{post} y \quad (14)$$

This corresponds to a static variant of HC in which $H^{pre}$ and $H^{post}$ are learned but do not depend on token-level input, unlike dynamic mappings described in the mHC paper's broader formulation.

### 2.4.2 Manifold-Constrained Residual Mixing

Residual mixing between streams is implemented by a learned matrix of logits $Z \in R^{nxn}$. Each forward pass produces a mixing matrix $H \in R^{nxn}$ via a Sinkhorn-style procedure that alternately normalizes rows and columns after exponentiation. This mirrors the Sinkhorn-Knopp projection described in the mHC paper.

Given $H$, the mixed residual stream is

$$X_i^{mixed} = \sum_{j=1}^{n} H_{ij} X_j \quad (15)$$

The final stream update for one layer is

$$X^{+} = X^{mixed} + \Delta X. \quad (16)$$

This corresponds to using a constrained residual mapping to mix streams while adding the scattered layer output. Conceptually, this is aligned with the mHC motivation of restoring stable identity-like behavior by constraining $H$ to a doubly stochastic manifold.

### 2.4.3 Output Aggregation and Language Modeling Head

After $L$ multi-stream blocks, we aggregate streams to a single hidden state using another learned simplex weight vector $w^{agg} \in \Delta^{n-1}$:

$$h_{out} = \sum_{i=1}^{n} w_i^{agg} X_i^{(L)} \tag{17}$$

A final normalization and tied output head produce logits and cross-entropy loss as in the baseline.

#### 2.4.4 Relationship to mHC Parameterization

It is important to clarify what is and is not implemented relative to the mHC paper. The mHC paper describes a parameterization in which mappings can be dynamically computed using projections of a normalized flattened hidden state, and then projected via sigmoid or Sinkhorn-Knopp to satisfy manifold constraints. Our implementation uses static learnable parameters for pre and post simplex weights and for residual mixing logits, and does not implement the dynamic mapping parameterization shown in the paper. This difference affects both expressivity and compute patterns, and it is one reason the efficiency claims in the mHC paper should not be transferred to our setting. The mHC paper's infrastructure section is explicit that specialized kernels and careful mixed precision design are used to reduce overhead in their large-scale implementation.

### 2.5 Stream-Specialized Adapters

We introduce stream-specialized adapters to add stream-specific capacity without changing the SSM core. The adapters are enabled optionally and controlled by a bottleneck rank $r$ and an adapter dropout rate.

The adapter mechanism uses a shared down projection $W_{\downarrow} \in R^{Dxr}$ and up projection $W_{\uparrow} \in R^{rxD}$ and per-stream scaling vectors. A SiLU nonlinearity is applied in the bottleneck. This design is conceptually aligned with the general adapter idea of inserting a compact bottleneck module, although our use differs from the task adaptation setting emphasized by prior adapter literature.

#### 2.5.1 Pre-Adapter

The pre-adapter operates on the stream tensor $X \in R^{BxTxnxD}$. A normalization is applied, then the down projection, activation, and per-stream scaling in bottleneck space. For stream $i$, let $\gamma_i^{pre} \in R^r$ be a learned scaling vector. The pre-adapter update can be expressed as

$$X_i \leftarrow X_i + Drop(W_\uparrow(\phi(W_\downarrow Norm(X_i)) \odot \gamma_i^{pre})) \tag{18}$$

Where $\phi$ is SiLU and $Drop$ is dropout.

### 2.5.2 Post-Adapter

The post-adapter operates on the single-stream SSM output $y \in R^{BxTxD}$ and produces a stream-wise tensor prior to scattering. Let $\gamma_i^{post} \in R^r$ be a learned scaling vector per stream. The update is

$$y_i = y + Drop(W_\uparrow(\phi(W_\downarrow Norm(y)) \odot \gamma_i^{post})) \tag{19}$$

The scattered update then use $y_i$ rather than a broadcast $y$, yielding

$$\Delta X_i = w_i^{post} y_i \tag{20}$$

This creates a direct mechanism for stream specialization both before aggregation into the SSM input and after the SSM produces an output, while sharing most adapter parameters.

## 2.6 Training Procedure

All models are trained for 10 epochs with sequence length $T = 256$, batch size $B = 16$, number of layers $L = 8$, and model dimension $D = 512$. Automatic mixed precision is enabled on CUDA using autocast and gradient scaling[21]. Optimization uses AdamW[22].

Gradient clipping is applied with a maximum norm of 1.0. Weight decay is set to 0.1, and the learning rate is 3e-4. These settings are fixed across compared variants to isolate the impact of the residual topology and adapters.

Validation evaluation is run periodically during training at fixed step intervals. At the end of training, a final validation loss and perplexity are reported. All experiments are done with NVIDIA GeForce GTX 1650 Ti GPU and an Intel Core i7-10750H CPU.

### 2.7 Benchmarking and Fair Checkpointing

A separate benchmarking script measures tokens per second by running a fixed number of optimizer steps on synthetic random token data, ensuring backward pass and optimizer updates are included. Because such steps modify weights, the benchmarking procedure saves a copy of the model state prior to timing and restores it afterward before performing perplexity evaluation. This produces a “fair” perplexity measurement from the original checkpoint after throughput timing. Peak GPU memory is measured via CUDA peak allocation statistics.

## 3. Results

This section reports both the stepwise training evaluation logs and the checkpoint-based fair benchmarking summary provided.

### 3.1 Stepwise Validation During Training

All three variants were trained under the same core configuration and evaluated every 500 steps. The baseline SSM training reported validation loss and perplexity at steps 500 through 5500, followed by a final evaluation after 10 epochs. The static mHC and mHC plus adapters variants reported evaluations at the same intervals except that the static mHC log does not display an evaluation at step 3000.

To present a complete and readable record, Figures 2 and 3, and Tables 1, 2 and 3 summarize the validation results reported in the logs during training for each model. The values in these tables are taken directly from the training logs.

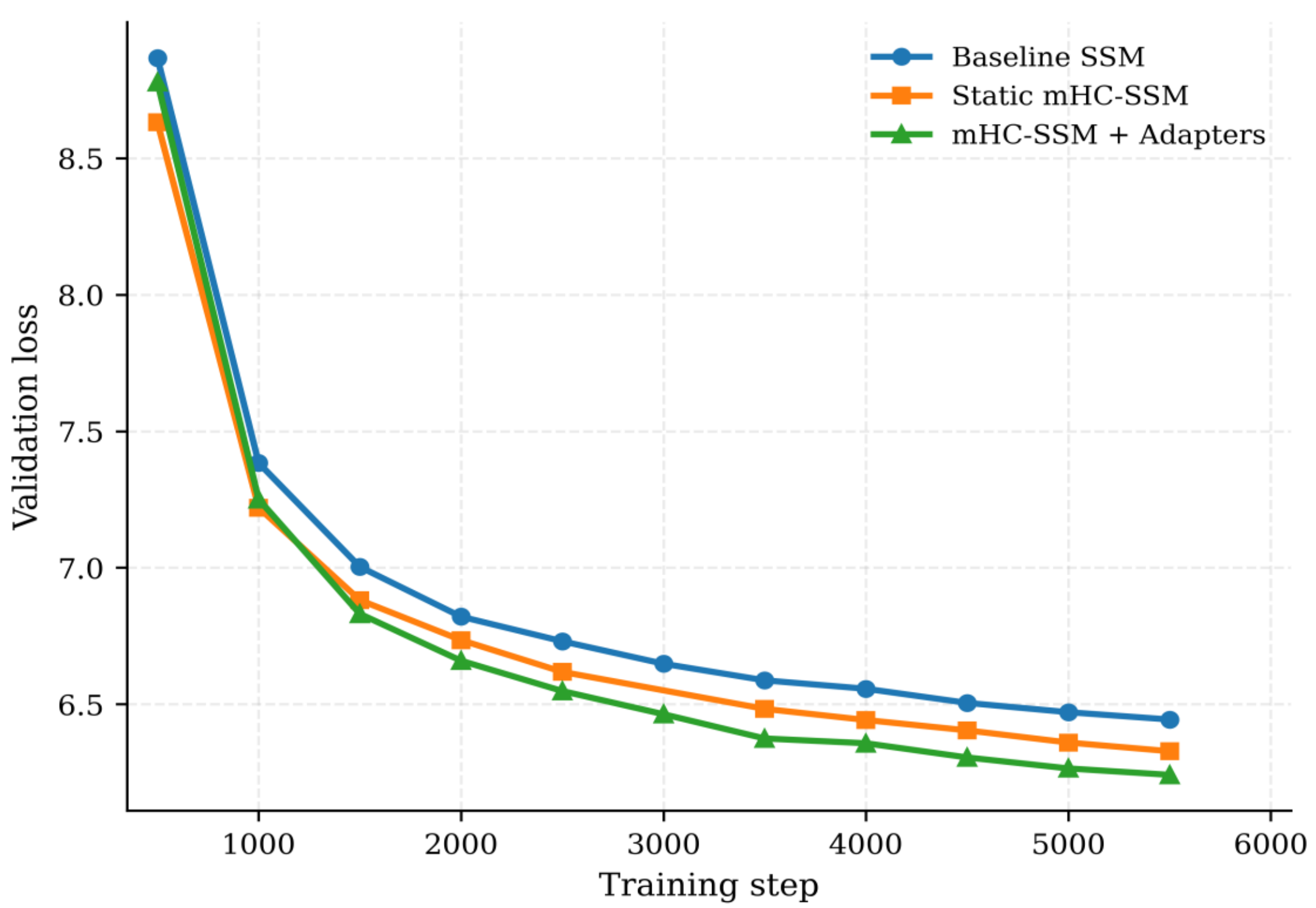

Figure 2 – Validation Loss During Training

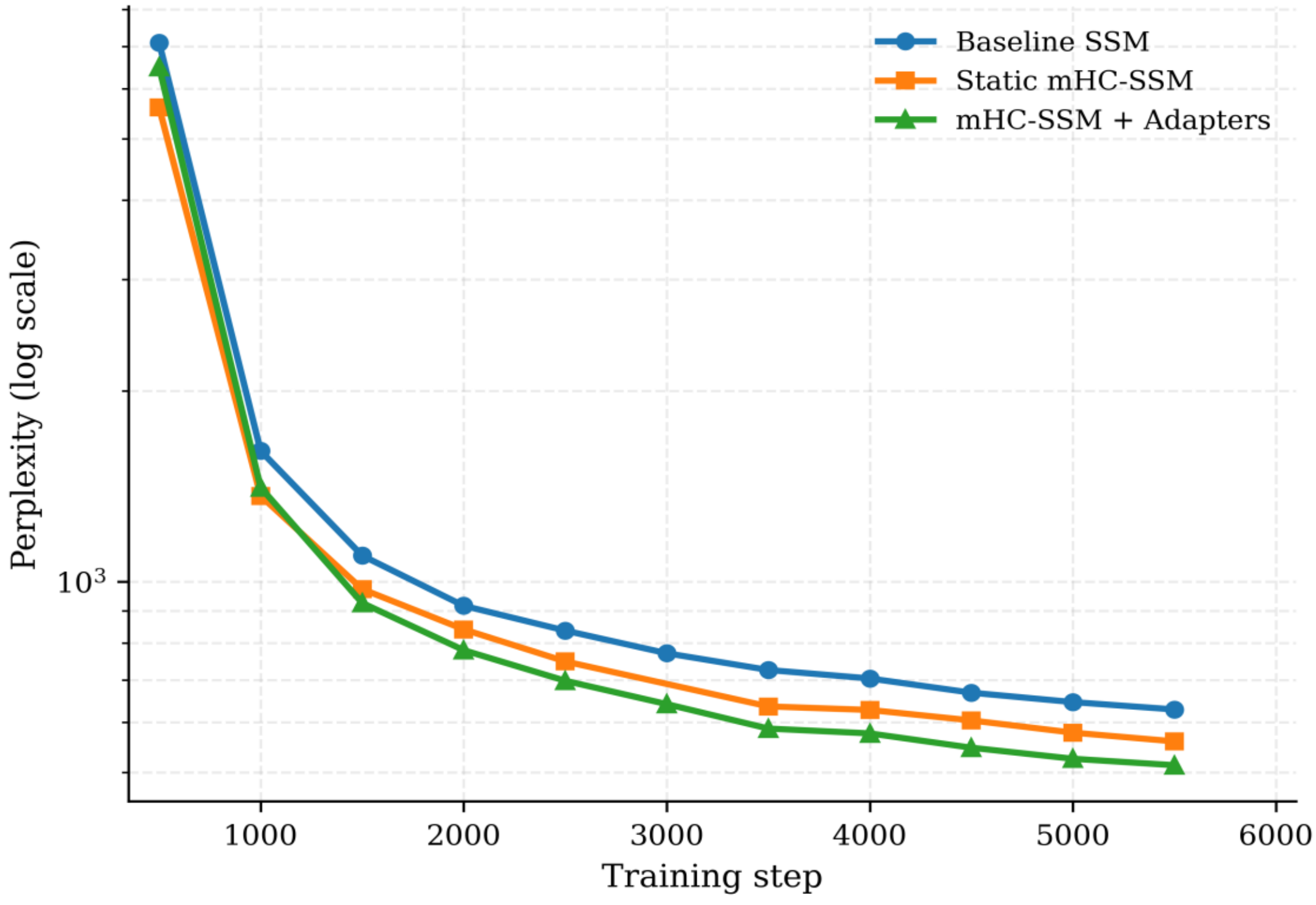

Figure 3 – Validation Perplexity During Training

Table 1 – Baseline SSM validation during training

| Step | Val Loss | PPL |
| --- | --- | --- |
| 500 | 8.8672 | 7095.36 |
| 1000 | 7.3836 | 1609.37 |
| 1500 | 7.0030 | 1099.95 |
| 2000 | 6.8205 | 916.46 |
| 2500 | 6.7304 | 837.50 |
| 3000 | 6.6480 | 771.21 |
| 3500 | 6.5873 | 725.82 |
| 4000 | 6.5564 | 703.74 |
| 4500 | 6.5047 | 668.28 |
| 5000 | 6.4706 | 645.85 |
| 5500 | 6.4438 | 628.78 |
| Final | 6.3964 | 599.68 |

Table 2 – Static mHC SSM validation during training

| Step | Val Loss | PPL |
| --- | --- | --- |
| 500 | 8.6316 | 5606.09 |
| 1000 | 7.2200 | 1366.45 |
| 1500 | 6.8812 | 973.81 |
| 2000 | 6.7348 | 841.20 |
| 2500 | 6.6183 | 748.65 |
| 3500 | 6.4828 | 635.77 |
| 4000 | 6.4420 | 627.64 |
| 4500 | 6.4040 | 604.25 |
| 5000 | 6.3595 | 577.97 |
| 5500 | 6.3277 | 559.89 |
| Final | 6.2889 | 538.54 |

Table 3 – mHC SSM plus stream adapters validation during training

| Step | Val Loss | PPL |
| --- | --- | --- |
| 500 | 8.7784 | 6492.66 |
| 1000 | 7.2504 | 1408.67 |
| 1500 | 6.8308 | 925.90 |
| 2000 | 6.6591 | 779.87 |
| 2500 | 6.5481 | 697.88 |
| 3000 | 6.4632 | 641.11 |
| 3500 | 6.3743 | 586.60 |
| 4000 | 6.3568 | 576.37 |
| 4500 | 6.3048 | 547.18 |
| 5000 | 6.2646 | 525.65 |
| 5500 | 6.2411 | 513.45 |
| Final | 6.1840 | 484.93 |

Several descriptive observations follow from these logs.

First, all models show a rapid decrease in perplexity during the early phase of training, followed by a slower improvement later. Second, at comparable steps, both multi-stream mHC variants typically report lower validation loss than the baseline. For example, at step 2000, baseline reports 6.8205, static mHC reports 6.7348, and mHC plus adapters reports 6.6591. Third, the adapter-augmented variant tends to report the lowest validation loss among the three configurations at later steps, culminating in a final loss of 6.1840 and perplexity of 484.93.

These trends are consistent with the idea that constrained stream mixing and stream-specific capacity can improve modeling quality under the chosen configuration, though the logs alone do not establish whether the differences are robust across random seeds or hyperparameter variations[23].

### 3.2 Fair Benchmark Results with Checkpoints

A separate checkpoint-based benchmarking summary is reported as follows.

Table 4 – Final benchmark results under checkpoint-based fair evaluation with throughput timing and weight restoration.

| Model | Val Loss | PPL | Tokens/sec | Peak Mem |
|---|---|---|---|---|
| Baseline SSM[8,9,14] | 6.3507 | 572.91 | 1025.52 | 2365 MB |
| Static mHC SSM[4,5] | 6.2448 | 515.35 | 964.81 | 2568 MB |
| mHC SSM + Adapters | 6.1353 | 461.88 | 938.90 | 3092 MB |

This table provides the primary comparison used in later discussion for quality and systems tradeoffs because it attempts to ensure perplexity is computed from checkpoint weights after timing.

Relative to baseline in this benchmark report, static mHC improves validation loss by 0.1059 and reduces perplexity from 572.91 to 515.35. The throughput decreases from 1025.52 tokens per second to 964.81 tokens per second, while peak memory increases from 2365 MB to 2568 MB.

Relative to static mHC, adding adapters further improves validation loss by 0.1095 and reduces perplexity from 515.35 to 461.88. The throughput decreases further to 938.90 tokens per second, and peak memory increases to 3092 MB.

### 3.3 Summary of Observed Tradeoffs

The results show a consistent pattern across the training logs and the benchmark summary: moving from baseline to static mHC improves validation loss and perplexity while increasing memory use and reducing throughput, and adding stream-specialized adapters yields further improvements in perplexity with additional memory increase and a modest additional throughput reduction.

## 4. Discussion

A central claim in the mHC paper is that unconstrained residual stream mixing in HC can lead to instability because the composite residual mapping across depth does not preserve global mean signal intensity across streams, enabling unbounded amplification or attenuation as depth increases. To restore identity-mapping-like stability in a multi-stream setting, the paper constrains the residual mixing matrix to be doubly stochastic by projecting onto the Birkhoff polytope. It then motivates this choice by stating that the spectral norm of a doubly stochastic matrix is bounded by one and that the set is closed under multiplication, implying that the composite mapping across layers remains doubly stochastic.

In the SSM setting evaluated here, the evidence is empirical rather than diagnostic. We did not measure gradients, spectral norms, or other stability proxies. Nonetheless, the results indicate that replacing a single-stream residual SSM model with a static multi-stream architecture that includes Sinkhorn-projected residual mixing can improve perplexity on WikiText-2 in this configuration. One interpretation is that stream-wise mixing provides an additional representational mechanism that complements the SSM block's sequence modeling behavior, and that constraining mixing helps keep training stable enough to realize these gains.

However, it should be emphasized that our implementation uses a static mapping, whereas the mHC paper's method includes dynamic mappings in its main formulation. Therefore, the quality improvements observed here are best viewed as evidence that the general idea of constrained multi-stream residual mixing can be beneficial even in a simplified static form, rather than as a direct validation of the full mHC system.

It is useful to consider what multi-stream residual topologies can offer in an SSM model.

A single-stream residual block computes an update in the same feature space as the residual path. In contrast, multi-stream representations allow the model to maintain parallel feature subspaces that can be mixed at each layer. The pre aggregation step allows the model to compute a single-stream representation used as the input to the SSM block, potentially emphasizing certain streams more than others. The post scattering step allows the SSM output to influence streams differentially.

The residual mixing matrix then creates a structured mechanism to blend information across streams. If this matrix is unconstrained, it can behave like an arbitrary linear transform over streams, which in principle offers high flexibility but might lead to unstable accumulation when composed repeatedly. The mHC paper's choice to constrain it to a doubly stochastic manifold is presented as a way to preserve stability while still allowing mutual interaction among streams.

In our implementation, pre and post weights are simplex-constrained by softmax, making them non-negative and summing to one. This is consistent with the mHC paper's emphasis on non-negativity constraints to prevent cancellation, though the paper uses a particular sigmoid-based parameterization for pre and post mappings in one formulation. These design choices collectively bias the system toward convex mixing rather than arbitrary signed combinations, which may help avoid destructive interference patterns.

The adapter-augmented variant shows additional improvement beyond static mHC in both training logs and the checkpoint-based benchmark. A plausible explanation is that stream-specialized adapters create a stronger incentive and mechanism for streams to represent different aspects of the sequence, because each stream has its own scaling parameters that modulate a shared bottleneck transformation. This provides a stream-specific degree of freedom that is separate from the global stream mixing weights.

This design is conceptually related to adapter bottleneck modules, which are known to provide expressive capacity with comparatively few parameters in transfer learning settings. Although our setting differs because we train from scratch rather than adapting a pretrained model, the general principle that small bottleneck modules can add useful capacity may still apply. Similarly, LoRA demonstrates that low-rank modifications can be effective for parameter-efficient adaptation. Our adapters are not implemented as low-rank updates to the same linear layers in the manner of LoRA, but the shared low-dimensional bottleneck and per-stream scaling share a similar structural motivation.

It is also possible that adapters improve optimization by providing a more direct route for gradient flow into per-stream parameters, or by adding a smoother adjustment mechanism around the SSM core. Such interpretations are speculative here because auxiliary diagnostics were not logged, and only end metrics are reported.

The checkpoint-based benchmark results quantify practical tradeoffs. Multi-stream architectures require storing and processing an expanded residual stream. Even when the inner block operates on a single stream, the expanded representation must be maintained for residual mixing and for expansion and aggregation operations. Consequently, tokens per second decrease and peak memory increases as we move from baseline to static mHC and then to mHC plus adapters.

This aligns with the mHC paper's argument that memory access overhead is a key bottleneck when widening residual streams. The mHC paper describes a series of infrastructure optimizations aimed at addressing these overheads, including kernel fusion, careful reordering of operations for mathematical equivalence, fused operations to reduce memory bandwidth bottlenecks, and custom kernels for Sinkhorn iterations and their backward pass. Our implementation does not incorporate such optimizations, so the throughput and memory measurements here reflect a straightforward PyTorch implementation and should not be used to infer what an optimized production kernel stack might achieve.

Nonetheless, even in this unoptimized setting, the throughput reductions are modest relative to the perplexity improvements reported. The baseline to static mHC change reduces throughput from 1025.52 to 964.81 tokens per second, and the adapter addition reduces it further to 938.90 tokens per second. Peak memory increases more noticeably, especially when adapters are enabled, rising from 2365 MB to 2568 MB to 3092 MB. These differences are operationally relevant for GPU-constrained training settings.

The mHC paper explicitly states three theoretical properties of doubly stochastic residual mixing that motivate stability: spectral norm bounded by one, closure under multiplication, and a geometric interpretation as a convex hull of permutations. In our implementation, the residual mixing matrix is produced by a Sinkhorn-style iterative normalization of exponentiated logits, matching the high-level procedure described in the paper. In the paper's large-scale experiments, $t_{max} = 20$ is used as practical iteration count. Our implementation uses fewer iterations for practical reasons, which may result in a matrix that is only approximately doubly stochastic.

Despite this approximation, the results suggest that even approximate doubly stochastic mixing can yield improvements. This could mean that exact convergence is not necessary for benefits in small-scale settings, or it could mean that other aspects of multi-stream structure dominate in this scale regime. Disentangling these effects would require controlled experiments varying Sinkhorn iterations and temperature while measuring both quality and stability proxies.

### 4.1. Limitations

This study has several limitations that bound what can be concluded.

First, evaluation is limited to WikiText-2 and a single configuration of model size, depth, stream count, and sequence length. It is unknown whether the same trends hold on other corpora or at larger scales.

Second, the results provided correspond to single runs for each configuration, and variance across random seeds is not characterized. Without multiple seeds, it is difficult to quantify the robustness of small loss differences.

Third, the implementation is a static variant of mHC-style mixing. The mHC paper describes dynamic mappings and a particular parameterization and projection scheme involving sigmoids for pre and post mappings and Sinkhorn-Knopp for residual mappings. Our design uses simplex-constrained softmax weights for pre and post mappings and a static residual mixing matrix projected via Sinkhorn-style normalization. Therefore, our results do not validate the full dynamic mHC method.

Fourth, the efficiency analysis is limited to PyTorch-level benchmarking. The mHC paper emphasizes that its scalability depends on specialized infrastructure, including fused kernels and custom backward passes. Since our code does not implement these kernels, the throughput results represent an upper bound on overhead in this unoptimized setting, and they cannot be directly compared to the paper's reported large-scale overhead outcomes.

Fifth, the study does not include direct stability diagnostics. The mHC paper motivates its constraints through stability properties of doubly stochastic matrices. We do not measure gradient norms, activation norms, or spectral properties, so the mechanism behind improvements remains inferential.

### 4.2. Practical implications.

For practitioners working with SSM-based language models, these results suggest that constrained multi-stream residual mixing can yield meaningful perplexity improvements even in a simplified static form. This is relevant because it implies that macro-architectural improvements inspired by mHC are not necessarily tied to attention-based backbones.

The results also clarify tradeoffs. Multi-stream residual designs and adapters increase peak GPU memory, which may become the primary constraint before compute throughput does, especially on consumer GPUs. When adapters are enabled, peak memory increases to approximately 3092 MB in the benchmark report, which may be acceptable for some training setups but might constrain batch size or sequence length in others.

In settings where throughput is critical, the observed reduction from baseline to mHC plus adapters is about 8 percent in tokens per second relative to baseline. Whether this tradeoff is worthwhile depends on target quality requirements and available memory headroom.

The mHC paper emphasizes that infrastructure optimization can reduce overhead. This suggests that in production contexts, it may be feasible to regain throughput by adopting fused kernels for normalization, mixing, and Sinkhorn projection. However, such engineering is beyond the scope of the present implementation.

### 4.3. Future Work

Several future directions follow directly from the gaps and limitations.

A first direction is to evaluate scaling trends. This includes testing larger model dimensions and deeper networks, increasing sequence length, and evaluating on additional language modeling datasets beyond WikiText-2. Such studies would help determine whether improvements persist or grow with scale.

A second direction is to explore the impact of Sinkhorn iteration count and temperature. The mHC paper uses $t_{max} = 20$ in its experiments. Our implementation uses fewer iterations. Controlled ablations varying iteration counts and measuring both approximate doubly stochasticity metrics and downstream perplexity could clarify how strongly exact manifold projection matters.

A third direction is to implement dynamic mappings more directly aligned with the mHC paper's parameterization, where mappings depend on the normalized hidden state and are projected through specific manifolds. This would allow a closer comparison to the method as introduced, and might yield additional improvements at the cost of compute.

A fourth direction is to add diagnostics and auxiliary objectives. For instance, one could log stream diversity statistics or other measures of stream differentiation during training. This would help test whether adapters truly encourage stream specialization.

A fifth direction is to explore alternative stream-specialization mechanisms. The current design uses per-stream scaling in a shared bottleneck. Other options include per-stream low-rank projections, per-stream gating, or constrained orthogonal transformations. Such variations could change the quality and systems tradeoff surface.

A final direction is systems engineering. The mHC paper describes kernel fusion and custom backward designs for mHC's projections and mixing. Implementing even a subset of these ideas in a PyTorch extension or CUDA kernel could provide a clearer picture of the true overhead-quality tradeoff when mHC is engineered for efficiency.

## 5. Conclusion

This paper studied an mHC-inspired, manifold-constrained hyper-connection style residual mixing mechanism in a state space language model setting. A baseline single-stream SSM language model was compared with a static multi-stream variant that applies Sinkhorn-projected residual mixing, and with an extension that adds stream-specialized adapters implemented via a shared bottleneck with per-stream scaling.

Across the provided training logs and a checkpoint-restored fair benchmark report, both static mHC mixing and mHC plus adapters improved validation loss and perplexity on WikiText-2 relative to baseline. The improvements were accompanied by an increase in peak GPU memory and a modest decrease in tokens per second throughput, consistent with the idea that widened residual streams introduce memory access overhead. These results suggest that mHC-inspired constrained multi-stream topologies can provide measurable benefits outside of attention-based backbones, and that lightweight stream-specialization mechanisms can further improve quality.